\documentclass[preprint,12pt]{elsarticle}

\usepackage{graphics}
\usepackage{mathptmx}
\usepackage{times}
\usepackage{amsmath}
\usepackage{amssymb}
\usepackage{setspace}

\newtheorem{Remark}{Remark}
\newtheorem{assump}{Assumption}
\newtheorem{Prop}{Proposition}

\newcommand{\pos}{\textbf{r}}
\newcommand{\C}{\textbf{c}}

\journal{IFAC Journal of Systems and Control}

\begin{document}

\begin{frontmatter}

\title{Cooperative Gradient Estimation for Multi-Robot Source Seeking with Heteroskedastic Measurements}

\author[label1]{Alessandro Renzaglia}
\author[label2]{Lara~Bri\~{n}\'{o}n-Arranz}
\affiliation[label1]{organization={Inria, INSA Lyon, CITI},
             postcode={69100},
             city={Villeurbanne},
             country={France}}

\affiliation[label2]{organization={Univ. Grenoble Alpes, CNRS, Grenoble INP, GIPSA-lab},
             postcode={38000},
             city={Grenoble},
             country={France}}

\begin{abstract}
This paper proposes a novel approach for localizing the source of a three-dimensional scalar field using a team of heterogeneous cooperative mobile robots. Each robot measures the field intensity at its location and shares the information within the team. Measurement noise is explicitly considered, and heterogeneity arises from robot-dependent sensing accuracies, leading to heteroskedastic data.
The proposed framework combines a weighted centroidal Voronoi tessellation-based formation strategy, defined on a spherical surface and parameterized by sensor noise variances, with a Weighted Least Squares estimator to compute the field gradient at the formation center. By explicitly accounting for sensor heterogeneity in both formation design and estimation, the method achieves robust gradient estimation under noisy measurements.
In particular, we show analytically that this construction is nearly optimal. The covariance of the gradient estimate is derived in closed form, providing a lower bound on the attainable estimation variance for any formation on the sphere, and the proposed tessellation is shown to asymptotically satisfy the conditions under which this bound is attained. The estimated gradient is then exploited within a gradient-ascent strategy to guide the team toward the source location.
Simulation results demonstrate the effectiveness of the proposed approach compared to alternative strategies and highlight its intrinsic ability to reconfigure online in response to robot failures, thereby enhancing its adaptability.

\end{abstract}

\begin{keyword}
Sensor Networks, Gradient Estimation, Multi-Robot Systems, Source Seeking
\end{keyword}

\end{frontmatter}

\section{Introduction}
\label{sec:intro}
Cooperative source seeking is a critical challenge in multi-robot applications due to its importance in various scenarios, including search and rescue missions, environmental monitoring, and pollution source detection \cite{bayat2017survey}, \cite{isler2015finding}, \cite{gas_survey}. In this context, an unknown source generates a spatially distributed signal field.
A team of autonomous robots, equipped with sensors capable of measuring the signal intensity at their locations, works together to localize the source by leveraging the collected information.
Despite being a widely studied problem in the literature, efficient and accurate source localization based only on sparse and noisy field intensity measurements remains a challenging problem. As discussed in more detail in the following section, most existing solutions focus on two-dimensional environments and often do not analyze the impact of measurement noise. When noise is considered, it is always assumed that the measurements collected by the robots have equal variance, i.e., the collected data are homoscedastic. This is not always the case in many real-world applications, where heterogeneous robots may be equipped with sensors of different quality and characteristics. For example, a few robots equipped with a high-quality, high-value sensor that provides very accurate measurements can be deployed together with robots equipped with cheaper, less accurate sensors.

In this paper, we translate the source localization task into the problem of accurately estimating the gradient of the generated scalar field from sparse and noisy measurements. Since the reliability of the collected heteroskedastic data varies across sensors, this heterogeneity can significantly affect the quality of the estimated gradient and, consequently, the accuracy of the final source localization. Therefore, such information must be taken into account not only in the estimator design but also during the robots’ planning phase. In particular, determining where each robot should collect its measurements becomes a key decision for maximizing the information gathered at the team level.
To this end, this work incorporates sensor heterogeneity directly into the design of the measurement locations and the coordination strategy. More precisely, we aim to answer the following question:
How should a heterogeneous team of robots be coordinated so that their sensing capabilities are exploited to achieve reliable configurations for efficient and robust source localization?

The main contribution of this paper is to present a new approach to address this issue and to perform the source search within a 3D signal field with a multi-robot system collecting heteroskedastic measurements.
The main novelty of this solution is to propose a coordinated measurement strategy based on multi-robot formations obtained via weighted centroidal Voronoi tessellations on spherical surfaces. The weights, which are related to the noise variance of each robot, allow the formation to incorporate the heteroskedasticity of the data and reflect it in the geometric properties of the multi-robot configuration. As a result, the formation presents a different density of robots depending on sensor accuracy, providing higher spatial coverage near less accurate, and therefore less reliable, sensors. The data collected from this formation are then used to obtain a gradient estimate using a weighted least squares method. Furthermore, we provide an analytical justification of the resulting design in terms of estimation accuracy. We derive a closed-form expression for the covariance of the gradient estimate and a corresponding lower bound on the estimation variance attainable by any formation on the spherical surface, together with the two geometric conditions under which this bound is reached. We then prove that the weighted tessellation satisfies both conditions, up to an approximation error that vanishes as the team size grows, and verify numerically that the resulting formation lies within a few percent of the bound already for small teams.
Additionally, a key feature of the proposed framework is its intrinsic reconfiguration capability. Changes in team composition or sensor variances are directly reflected in the weighting function defining the CVT, leading to a modified set of target measurement locations. As a result, the sensing formation is automatically reshaped through the formation controller without requiring explicit fault handling or redesign.

\subsection{Related Work}
\label{sec:rel_work}

Source seeking with autonomous robots has been extensively studied in recent years, with numerous solutions proposed for both single and multi-robot systems. However, the majority of these approaches are limited to 2D search spaces, while solutions for 3D environments are still scarce. Existing solutions can be broadly categorized into two classes: gradient-free and gradient-based approaches.

Gradient-free methods have gained popularity due to their simplicity and robustness in environments where gradient information is difficult to obtain, for instance, due to discontinuities in the signal.
In this context, a significant body of work has focused on Particle Swarm Optimization (PSO) algorithms. For instance, \cite{zou2015particle} and \cite{zhang2023pso} proposed PSO-based strategies for efficient source localization in 2D environments.
\cite{li2022multi} introduced an adapted version of the Quantum Particle Swarm Optimization (QPSO) algorithm, which proposed leading-following behaviors in a swarm-based approach to improve search efficiency in 2D spaces.
PSO-based solutions, however, can become unstable when individual field measurements are unreliable due to high levels of noise, especially with small-sized multi-robot teams. Single outliers can indeed corrupt personal and global best positions, misleading the entire team and potentially causing the optimization to stagnate or induce significant oscillations.
Other gradient-free approaches draw inspiration from biological systems. A method inspired by schools of fish, where robots adjust their speed based on environmental cues to locate darker areas, is presented in \cite{Wu2015}. Similarly, in \cite{tran2023multi}, a novel flocking control strategy is proposed for multi-robot exploration and gas field mapping within a 2D framework.
Building on this idea, \cite{tran2024collaborative} presents a collaborative gas source localization strategy employing networked nano-drones operating in unknown and cluttered environments, achieved through hybrid integration of Adaptive Robotic Particle Swarm Optimization (ARPSO) with Bidirectional Brain Emotional Learning (BBEL).
The use of multi-robot formations is then treated in \cite{du2024simultaneous}, which proposed a gradient-free approach addressing source localization and formation control for ground robots.
Communication can also be a crucial aspect in multi-robot strategies, and \cite{al2018gradient} introduced a consensus-like perception algorithm that allows estimating the source location without explicit communication.

The second class of solutions leverages an explicit gradient estimation of the sensed field to guide robots toward the source. A key challenge in these approaches is obtaining a reliable estimation from sparse and noisy measurements. To do so, two main, often interconnected, problems usually arise: \textit{i)} how to get the most accurate estimate of the gradient given a set of sparse noisy measurements at a given time, and \textit{ii)} where to locate the robots to obtain the set of measurements that will provide the best gradient approximation, given the estimation method. Several techniques have been proposed to address this issue, including consensus-based algorithms \cite{fabbiano2014source}, cooperative Kalman filters \cite{Wu2011}, and Least Squares methods \cite{ogren2004}. However, these solutions often focus on the estimation problem, overlooking the study and optimization of sensor placement, or present approaches whose use is limited by strict requirements on specific robot formations, which are only defined for a limited number of robots, as in \cite{fabbiano2014source}, hence reducing their applicability and robustness to robot failures.

To address these limitations, some works have proposed multi-robot formations with specific geometric constraints to ensure reliable gradient estimation with proved guarantees \cite{marzat2015cooperative, Lara2019TRO}. In \cite{acuaviva2024cdc}, the authors showed that more generic swarm configurations can be adopted to estimate an ascending direction instead of the exact gradient.
While most of these works focus on 2D environments, a few have explored 3D scenarios.
In \cite{Lara2019TRO}, we introduced 3D symmetric multi-robot formations that enable the estimation of both the gradient and the Hessian of the field using simple averages of robot measurements and their relative positions. However, this approach requires an even number of robots and a centralized architecture. In a more recent work \cite{brinon2024voronoi}, we overcome these limitations by proposing a new class of formations based on Voronoi partitions over a spherical surface, enabling more flexible and scalable solutions. However, this solution remains suitable only for homogeneous sensing noise.

A third, more recent line of work replaces the explicit estimation step with a learned one. Reinforcement learning has been applied to derive source-seeking policies directly from sequences of measurements \cite{li2023aidrl}, while Gaussian-process regression and Bayesian optimization have been used to build a probabilistic model of the field and to select informative sampling locations \cite{coleman2023exploration, huang2025distributedgp}. Neural-network field approximation has also been combined with formation control for multi-robot source localization \cite{li2025neuralformation}, although this line of work assumes a deterministic field with bounded approximation error rather than a stochastic, heteroskedastic measurement model.
These approaches are attractive precisely when the structure of the field or the statistics of the measurement noise are unknown a priori, since they do not require an explicit model of either. The counterpart is the need for training data or for an exploration phase, and the loss of the closed-form error characterization that a model-based estimator provides. The approach proposed here represents a complementary situation, where the noise statistics are assumed known, and this knowledge is exploited to obtain both an analytically characterized estimator and a formation design that is provably close to optimal.

The work presented in this paper represents, to the best of our knowledge, the first solution that explicitly considers a heterogeneous system collecting heteroskedastic measurements in a 3D environment and proposes a novel integration of a robust gradient estimation scheme with a formation control strategy that adapts to the differences in measurement noise variance.

\section{Methods}
\label{sec:methods}

\subsection{Problem Formulation}
\label{sec:problem}

Let us consider a scalar field $\sigma(\pos):\mathbb{R}^3 \rightarrow \mathbb{R}^+$ representing the intensity of a physical signal in $\pos$. The field reaches its maximum intensity at its \emph{unique} source $\pos^*$, and we assume this location to be initially unknown.
Denoting the signal gradient at a location $\pos$ as $\nabla\sigma(\pos)$, we consider the following additional assumption on the signal:
\begin{assump}\label{assum:signal3D}
The function $\sigma: \mathbb{R}^3 \rightarrow \mathbb{R}^+$ is continuously differentiable, i.e., $\sigma \in \mathcal{C}^1$, and its gradient is globally bounded. Moreover, $\nabla\sigma(\pos^*) = 0$ and $\nabla\sigma(\pos) \neq 0, \, \forall\, \pos \neq \pos^*$.
\end{assump}
A function $\sigma(\textbf{r})$ satisfying this assumption allows representing a wide variety of spatial models that are frequently used in existing source-seeking strategies.

In this context, we then assume having a formation of $N$ mobile robots of positions $\mathbf{R}=\{\pos_i\}$ with $\pos_i \in \mathbb{R}^3$ and $i=1, \ldots, N$ expressed in the inertial global frame.
The $N$ robots can be controlled to stabilize their position and velocity to a desired reference. As widely discussed in the robot control-related literature, their non-linear dynamics can be linearized or simplified to leverage standard control techniques (e.g., see \cite{Nascimento2019} for a survey) and ensure the correct tracking of a desired position.
In this work, the desired position $\pos_i^{ref}$ for each robot $i$ is defined by a multi-robot formation generator.

The robots have access to their positions in a global reference frame and are equipped with sensors able to collect measurements of the intensity of the signal field of interest. Considering these measurements as affected by Gaussian zero-mean white noise, we can express them as:
\begin{equation} \label{eq:measurements}
y_i(\pos_i) = \sigma(\pos_i)+\mu_i, \ \ \ \mu_i\sim\mathcal{N}(0,\nu_i^2)
\end{equation}
where $\nu_i^2$ represents the variance of the noise for robot $i$. In this work, we focus on scenarios where the variances $\nu_i^2$ are not the same for all robots. With this assumption, we can model a broad range of applications where different robots are equipped with sensors of the same type but from different manufacturers or with varying specifications. This heterogeneity can occur either by design, for instance, combining a few high-cost, high-accuracy sensors with more affordable, lower-precision ones, such as a laboratory-grade illuminance meter alongside basic photodiodes, or out of necessity due to hardware availability or constraints.
Sensors with different specifications usually present different noise characteristics, which can significantly influence the performance of estimation algorithms and affect the efficiency of the path planning and source localization strategy.
Beyond this, the model in eq.~\eqref{eq:measurements} is agnostic to the physical entity, requiring only the noise to be additive and a variance that is stable over the operating range and known in advance. Ambient temperature, light intensity, relative humidity, acoustic intensity, and gas concentration all have sensors of this kind (e.g., thermistors, photodiodes, capacitive hygrometers, microphones, or infrared gas detectors) whose noise is reported directly in the manufacturer specifications. The variances $\nu_i^2$ are thus obtained either from those specifications or from a pre-mission calibration in which each sensor is exposed to a known reference stimulus, and its sample variance is recorded.

The final goal of the multi-robot formation is to share the collected measurements and cooperatively exploit them to navigate the environment and eventually find and locate the source $\pos^*$ of the scalar field.

\subsection{Voronoi-based Gradient Estimation}
\label{sec:method}

A standard approach to locate the signal source based on noisy measurements of its intensity is to first cooperatively estimate the gradient of the field at a given point $\C$. The gradient estimate, denoted by $\widehat\nabla\sigma(\C)$, thanks to Assumption~\ref{assum:signal3D}, can indeed be used to guide the multi-robot formation via a classic gradient-ascent (GA) method \cite{ogren2004}, \cite{zhu2014}, \cite{marzat2015cooperative}.
By estimating the gradient at the center of our multi-robot formation, we can define its motion as:
\begin{equation} \label{eq:gradient_ascent}
\C(k+1)=\C(k)+\epsilon\,\widehat\nabla\sigma(\C(k))
\end{equation}
where $\C(k)\in\mathbb{R}^3$ represents the formation center at iteration $k$ and $\epsilon > 0$ is the step size of the GA algorithm.

It could be argued that, in real-world scenarios, Assumption \ref{assum:signal3D} might not always be valid, and local irregularities in the field could prevent the existence of a well-defined gradient at all times. However, these local effects can be mitigated, for instance, with the inclusion of an inertia term that could allow the formation center to maintain its motion in the absence of a gradient estimate, whether due to missing measurements or abrupt fluctuations. This inertia, or an exploratory motion as done in \cite{renzaglia2020search}, would help overcome short-term instabilities until new, stable gradient information becomes available.

In this section, we present the proposed approaches to obtain a robust gradient estimate $\widehat\nabla\sigma$ given a set of sparse heteroskedastic measurements and to define suitable multi-robot formations to collect the required data from suitable locations that take into account the heterogeneity in the sensors' noise.

\subsubsection{Gradient estimate}
\label{subsec:LS}

Let us consider the first-order Taylor expansion of the signal field about the point $\C$. The intensity of the field at robot's position $\pos_i$ for $i=1,\ldots,N$ can be expressed as:
\begin{equation} \label{eq:taylor}
\sigma(\pos_i) = \sigma(\C)+\nabla\sigma(\C)^\top\widetilde{\pos}_i+\varphi(\pos_i,\C),
\end{equation}
where $\widetilde{\pos}_i=\pos_i-\C$ and $\varphi(\pos_i,\C)$ denotes the remainder of the Taylor expansion. By exploiting the $N$ noisy measurements
$$\mathbf{y}=[y_1(\pos_1),...,y_N(\pos_N)]^\top,$$
as defined in eq.~\eqref{eq:measurements}, the expression in eq.~\eqref{eq:taylor} can be used to estimate the vector $\Gamma$:
\begin{equation}
    \Gamma = \begin{bmatrix} \sigma(\C) \\ \nabla\sigma(\C) \end{bmatrix} \,.
\end{equation}
As shown in literature \cite{ogren2004}, this can be formulated as an ordinary Least Squares (LS) problem where the objective is to minimize $||\mathbf{y}-\mathbf{P}\,\Gamma||^2$, with the matrix $\mathbf{P}$ being:
\begin{equation}
    \mathbf{P}= \begin{bmatrix} 1 & \tilde\pos_1^\top \\ \vdots & \vdots \\ 1 & \tilde\pos_N^\top\end{bmatrix} \,.
\end{equation}
The solution to this problem is given by
\begin{equation}
    \widehat \Gamma = (\mathbf{P}^\top\mathbf{P})^{-1}\mathbf{P}^\top\mathbf{y} \,.
\end{equation}

Ordinary LS corresponds to the maximum likelihood estimation when the noise term in the measurements $\mathbf{y}$ can be characterized as independent and identically distributed Gaussian white noise. This condition, defined by a constant noise variance, is referred to as homoskedasticity and represents the standard assumption in many analyses. In other words, the measurement noise is a Gaussian white noise with the same standard deviation $\nu$, i.e., $\mu_i\sim\mathcal{N}(0,\nu^2), \, \forall i$.

However, in heterogeneous systems, such as the ones considered in this work, the noise variance may vary from robot to robot, resulting in heteroskedasticity within the data, i.e., $\mu_i\sim\mathcal{N}(0,\nu_i^2)$. In this case, we can instead minimize the weighted sum of squares $||\mathbf{W}^{1/2}(\mathbf{y}-\mathbf{P}\,\Gamma)||^2$ where $\mathbf{W}$ is a diagonal matrix with diagonal components equal to the inverse of the variance, i.e., $W_{ii}= 1/\nu_i^2$. The solution to this Weighted Least Squares (WLS) problem becomes:
\begin{equation} \label{eq:WLS}
    \widehat\Gamma = (\mathbf{P}^\top\mathbf{W}\mathbf{P})^{-1}\mathbf{P}^\top\mathbf{W}\mathbf{y} \,.
\end{equation}
By using eq. \eqref{eq:WLS}, the estimation will rely less on measurements with a higher variance while trusting more the data associated with more accurate measurements, resulting in a better and more robust estimation.

\subsubsection{Weighted Voronoi-based Multi-Robot Formation Generator}
\label{subsec:voronoi}

\begin{figure}[t]
     \centering
    \includegraphics[width=.85\columnwidth]{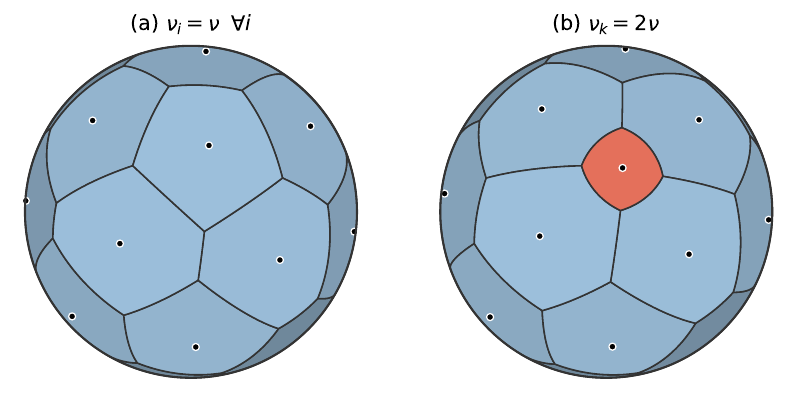}
        \caption{Two illustrative examples of CVT on a spherical surface, where the black dots represent the centroids of the Voronoi regions and the black lines their boundaries. Left: standard uniform CVT. Right: one robot has higher sensor noise, resulting in a smaller Voronoi region (red colored), forcing its neighbors to reduce their distance.}
        \label{fig:WCVT_illustration}
\end{figure}

The following problem to solve is how robots decide where to collect measurements to feed the gradient estimation module. In our scenario, this is equivalent to deciding how to define the multi-robot formation. Several works in the literature studied optimal $N$-robot configurations, trying to minimize the estimation error. Even though an optimal solution is, in general, too hard to obtain, especially in 3D dimensions, it has been shown that uniformly distributed formations with robots placed at a fixed distance from a formation center are suitable solutions to obtain robust estimates, such as in \cite{zhu2014}, \cite{Lara2019TRO}, \cite{fabbiano2014source}. However, most of these solutions either work only for a specific number of robots or are built on heuristics that obtain a good spatial distribution, but without being based on any formally defined criteria to optimize. An alternative way to achieve a proven spatially-balanced spread of robots is to formulate the problem as a cooperative coverage problem on a spherical surface $\mathcal{S}$, defined as
$$
\mathcal{S}(\C, D)=\{\mathbf{q} \in \mathbb{R}^3 \;\text{s.t.}\; ||\mathbf{q}-\C||=D\}\,.
$$
In our context, this ensures having the signal measurements taken from uniformly distributed locations around a reference point, which is crucial to obtain a good local representation of the scalar field and thus a reliable gradient estimation. In the cooperative coverage problem \cite{cortes2004coverage}, the optimization function $J$ to minimize is usually defined as:
\begin{equation} \label{eq:coverage_J}
J(\mathbf{R},\mathcal{A}) = \sum_i \int_{A_i} f(||\mathbf{q}-\mathbf{r}_i||)\,d\mathbf{q}
\end{equation}
where $\mathcal{A}$ is a partition of $\mathcal{S}$, with each robot $\mathbf{r}_i$ considered as responsible for its area of dominance $A_i$, and the coverage level function $f$ is most commonly assumed to be
$$f(||\mathbf{q}-\mathbf{r}_i||)=||\mathbf{q}-\mathbf{r}_i||^2.$$
To minimize $J$, it can first be observed that the optimal partition $\mathcal{A}$ for a fixed robots' configuration $\mathbf{R}$ is the Voronoi partition \cite{cortes2004coverage}. Voronoi partition, or tessellation, is a fundamental concept in Locational Optimization theory \cite{okabe1992} often used to design multi-robot strategies, where the partition of a bounded environment $\Omega \in \mathbb{R}^n$ generated by a set of points $\{\mathbf{r}_i\}_{i=1}^N$ is composed of $N$ regions $V_i$, defined as:
\begin{equation}
 V_i = \{\mathbf{q}\in\Omega : ||\mathbf{q}-\mathbf{r}_i|| \leq ||\mathbf{q}-\mathbf{r}_j|| \; \forall j\neq i\},\, i=1,\ldots,N \,.
\end{equation}
It can then be proved that the generators $\mathbf{r}_i^*$ corresponding to a {\it local} optimum of the coverage optimization problem are the centroids $C_{V_i}$ of the corresponding Voronoi regions, i.e.,
\begin{equation}
\mathbf{r}_i^* = C_{V_i} = \frac{\int_{V_i}\mathbf{q}\,d\mathbf{q}}{\int_{V_i}d\mathbf{q}} \quad \text{for}\;\;i=1,\ldots,N \,.
\end{equation}
The obtained tessellation, where all the generators correspond to the centroids of their regions, is called Centroidal Voronoi Tessellation (CVT).
Based on this idea, we recently proposed in \cite{brinon2024voronoi} a general class of formations where the robots' positions represent the generators of a CVT on a spherical surface of radius $D$. However, the analytic estimation of the gradient proposed in \cite{brinon2024voronoi} does not allow taking into account heterogeneity in the sensors' noise model. To do so, we introduce the information on the noise variance in the coverage distance function $f$ by defining $\tilde f_i$ as:
\begin{equation} \label{eq:f_tilde}
\tilde f_i(||\mathbf{q}-\mathbf{r}_i||)=\nu_i^2\;||\mathbf{q}-\mathbf{r}_i||^2\,.
\end{equation}
To cope with this modified version of the coverage optimization function, we can still rely on Voronoi tessellations, but now consider weighted distances. In most of the applications where Voronoi regions are used, the typical choice for the distance function is indeed the standard unweighted Euclidean distance, i.e., $d(\mathbf{x},\mathbf{y})=||(\mathbf{x}-\mathbf{y})||$. However, other solutions are possible, such as non-Euclidean distances or weighted distances, each of which results in different properties of the corresponding Voronoi regions.
In our case, to incorporate the effect of the $\tilde f_i$ functions, we introduce the following sensor-dependent distance function
\begin{equation} \label{eq:weighted_dist}
    d_i(\mathbf{x},\mathbf{y})=\nu_i \,||(\mathbf{x}-\mathbf{y})|| \,.
\end{equation}
The associated new Voronoi regions are thus defined as:
\begin{equation} \label{eq:weighted_voronoi}
\tilde V_i = \{\mathbf{q}\in\Omega : \nu_i \,||\mathbf{q}-\mathbf{r}_i|| \leq \nu_j \,||\mathbf{q}-\mathbf{r}_j|| \; \forall j\neq i\},\, i=1,\ldots,N \,.
\end{equation}
By adopting this new definition, the size of the region belonging to robot $i$ will continuously decrease as $\nu_i$ increases. As a result, in our proposed Weighted-CVT configuration, the density of robots over the spherical surface will not be uniform as in the unweighted case, but it will increase around a less accurate sensor, as shown in Fig.~\ref{fig:WCVT_illustration}. This will ensure having closer measurements around less reliable, and thus less trusted, robots and, on the contrary, sparser ones in regions covered by very reliable sensors.
It is worth noticing that, in the extreme case of $\nu_i\to\infty$ we will have $V_i\to\emptyset$, forcing the other robots to configure in a formation corresponding to the solution for the $N-1$ robots case.

To obtain a Weighted-CVT starting from an arbitrary set of generators $\{\mathbf{p}_i\}_{i=1}^N$, we adopt Lloyd's algorithm \cite{lloyd1982}. The main steps of this algorithm are: calculate the Voronoi partitions and the relative centers of mass corresponding to the current generators' configuration; move each generator toward the center of mass of its region; repeat this procedure until either convergence or the maximum number of iterations is reached. Although obtaining Voronoi regions, and consequently CVTs, is quite straightforward in 2D, this is not the case for general surfaces in 3D \cite{du2003voronoi}. To obtain a solution for our case, we firstly discretize our environment by creating a regular mesh of the spherical surface, following the procedure presented in \cite{mesh_sphere}\footnote{Note that this approach does not work for any number of points and is not suitable for small numbers of them, so it cannot be directly applied to create the desired multi-robot formation.}, to easily compute the Voronoi regions at each iteration. Then, a non-trivial step of implementing Lloyd's algorithm for a surface in 3D is the computation of the constrained mass centroid $\mathbf{p}_i^c$ of each region $i$. The classical unconstrained mass centroids $\mathbf{p}_i^*$ are indeed always inside the sphere and a projection on the surface is needed before moving each generator to its new position. However, for a spherical surface, it can be shown that $\mathbf{p}_i^c$ is simply represented by the intersection of the radial line passing through $\mathbf{p}_i^*$ with the spherical surface \cite{du2003voronoi}.
Fig.~\ref{fig:WCVT_illustration} shows an instance of a constrained CVT on a spherical surface with $20$ generators for both the unweighted and weighted scenarios. The obtained constrained CVT is then the basis for designing the multi-robot formation to perform our source-seeking task.

\subsubsection{Optimality Analysis of the Weighted-CVT Formation}
\label{subsec:optimality}

The formation generator introduced in the previous section was motivated by a field-coverage argument: robots with noisier sensors are assigned to smaller regions, so that the measurement density increases around them. We now analyze this construction from a more formal point of view to obtain a stronger theoretical justification. Specifically, we characterize in closed form the covariance of the WLS gradient estimate of eq.~\eqref{eq:WLS}, derive a lower bound on the estimation variance attainable by \emph{any} formation on the sphere, and show that the Weighted-CVT formation approaches this bound.

Consider the team of robots placed on the spherical surface $\mathcal{S}(\C,D)$, such that
\begin{equation}\label{eq:CVT_ideal_positions}
    \pos_i=\C+D\,\textbf{p}^c_i
\end{equation}
where $\textbf{p}^c_i$ are the constrained centroids of the Weighted-CVT with $\|\textbf{p}^c_i\|=1$. We neglect for the moment the Taylor remainder $\varphi(\pos_i,\C)$ in eq.~\eqref{eq:taylor}. Since the weighting matrix $\mathbf{W}$ is exactly the inverse of the covariance of the measurement noise, the estimator in eq.~\eqref{eq:WLS} is the best linear unbiased estimator of $\Gamma$, and its covariance carries no unknown scale factor:
\begin{equation} \label{eq:cov_gamma}
\mathrm{Cov}(\widehat\Gamma) = (\mathbf{P}^\top\mathbf{W}\mathbf{P})^{-1} \,.
\end{equation}
Only the lower-right $3\times3$ block of eq.~\eqref{eq:cov_gamma} is of interest, since the source-seeking law of eq.~\eqref{eq:gradient_ascent} uses $\widehat\nabla\sigma(\C)$ alone. Introducing the \emph{precision weights} and the total precision of the team,
\begin{equation} \label{eq:precision}
w_i \triangleq W_{ii} = \nu_i^{-2}\,, \qquad W_{tot} \triangleq \sum_{i=1}^N w_i = \sum_{i=1}^N \nu_i^{-2}\,,
\end{equation}
and according to eq.~\eqref{eq:CVT_ideal_positions}, the information matrix can be decomposed as
\begin{equation} \label{eq:infomatrix}
\mathbf{M} \triangleq \mathbf{P}^\top\mathbf{W}\mathbf{P} =
\begin{bmatrix}
W_{tot} & D\left(\sum_i w_i\textbf{p}^c_i\right)^\top \\[4pt]
D\sum_i w_i\textbf{p}^c_i & D^2\sum_i w_i\textbf{p}^c_i{\textbf{p}^c_i}^\top
\end{bmatrix} \,.
\end{equation}

\begin{Prop} \label{prop:cov}
Let $\bar{\textbf{p}}_w \triangleq \left(\sum_i w_i \textbf{p}^c_i\right)/W_{tot}$ denote the precision-weighted centroid of the robots' directions. Then
\begin{equation} \label{eq:cov_grad}
\mathrm{Cov}\left(\widehat\nabla\sigma(\C)\right) = \mathbf{S}_w^{-1}\,, \quad
\mathbf{S}_w = D^2\sum_{i=1}^N w_i\left(\textbf{p}^c_i-\bar{\textbf{p}}_w\right)\left(\textbf{p}^c_i-\bar{\textbf{p}}_w\right)^\top .
\end{equation}
\end{Prop}

\noindent\textit{Proof.} The gradient block of $\mathbf{M}^{-1}$ is the inverse of the Schur complement of the $(1,1)$ entry of eq.~\eqref{eq:infomatrix}, that is
$\mathbf{S}_w = D^2\left[\sum_i w_i\textbf{p}^c_i{\textbf{p}^c_i}^\top - W_{tot}\,\bar{\textbf{p}}_w\bar{\textbf{p}}_w^\top\right]$.
Expanding $\sum_i w_i(\textbf{p}^c_i-\bar{\textbf{p}}_w)(\textbf{p}^c_i-\bar{\textbf{p}}_w)^\top = \sum_i w_i\textbf{p}^c_i{\textbf{p}^c_i}^\top - W_{tot}\bar{\textbf{p}}_w\bar{\textbf{p}}_w^\top$ gives eq.~\eqref{eq:cov_grad}. $\blacksquare$

\vspace{.1cm}
Proposition~\ref{prop:cov} shows that the information available about the gradient is exactly $D^2$ times the precision-weighted scatter of the robots' directions about their precision-weighted centroid.
This has the important implication that the formation radius enters only through the factor $D^2$, so that larger formations are statistically preferable. This effect is limited only by the neglected truncation term $\varphi$ of the Taylor expansion in eq.~\eqref{eq:taylor}.

Since the robots lie on the sphere, we have $\|\textbf{p}^c_i\|=1$ for all $i$, and thus:
\begin{equation} \label{eq:trace_fixed}
\mathrm{tr}\left(\sum_i w_i\textbf{p}^c_i{\textbf{p}^c_i}^\top\right) = \sum_i w_i\|\textbf{p}^c_i\|^2 = W_{tot}\,,
\end{equation}
so that the total second moment depends only on the noise variances, not on where the robots are placed. Substituting eq.~\eqref{eq:trace_fixed} into eq.~\eqref{eq:cov_grad} gives $\mathrm{tr}\,\mathbf{S}_w = D^2 \, W_{tot}\left(1-\|\bar{\textbf{p}}_w\|^2\right)$. The design problem therefore splits in two: \textit{i)} the centroid term $\|\bar{\textbf{p}}_w\|$ sets how much information survives, and \textit{ii)} the eigenvalues of $\mathbf{S}_w$ set how it is shared among the three spatial directions.

\begin{Prop} \label{prop:bound}
For any configuration $\{\textbf{p}^c_i\}$ on the unit sphere, the following condition holds
\begin{equation} \label{eq:bound}
\mathrm{tr}\,\mathrm{Cov}\left(\widehat\nabla\sigma(\C)\right) \;\geq\; \frac{9}{D^2\sum_{i=1}^N \nu_i^{-2}}\;=\;\frac{9}{D^2 W_{tot}}\,,
\end{equation}
with equality if and only if the two following conditions hold simultaneously:
\begin{align}
\textit{(i)} && \sum_{i=1}^N w_i\,\textbf{p}^c_i &= \mathbf{0}\,, \label{eq:cond1}\\[2pt]
\textit{(ii)} && \sum_{i=1}^N w_i\,\textbf{p}^c_i{\textbf{p}^c_i}^\top &= \frac{W_{tot}}{3}\,\mathbf{I}_{3\times3}\,. \label{eq:cond2}
\end{align}
where $\mathbf{I}_{3\times3}$ denotes the identity matrix of appropriate dimensions.
\end{Prop}

\noindent\textit{Proof.} Let $\lambda_1,\lambda_2,\lambda_3>0$ be the eigenvalues of $\mathbf{S}_w$. From eq.~\eqref{eq:trace_fixed}, $\sum_j\lambda_j = D^2\, W_{tot}(1-\|\bar{\textbf{p}}_w\|^2)\leq D^2W_{tot}$, with equality if and only if $\bar{\textbf{p}}_w=\mathbf{0}$, which implies eq.~\eqref{eq:cond1}. By eq.~\eqref{eq:cov_grad}, the quantity to minimize is $\mathrm{tr}\,\mathbf{S}_w^{-1}=\sum_j\lambda_j^{-1}$, and the arithmetic--harmonic mean inequality gives $\sum_j \lambda_j^{-1}\geq 9/\sum_j\lambda_j$, with equality if and only if the three eigenvalues are equal. Substituting the maximal value $\sum_j\lambda_j = D^2W_{tot}$ yields eq.~\eqref{eq:bound}, and equal eigenvalues together with $\bar{\textbf{p}}_w=\mathbf{0}$ is precisely eq.~\eqref{eq:cond2}. $\blacksquare$

\begin{Remark}
The bound in eq.~\eqref{eq:bound} depends on the sensors only through the total precision $W_{tot}=\sum_i\nu_i^{-2}$, and not on how that precision is distributed across the team. This observation has the important implication that discarding the least accurate robots in the team is necessarily suboptimal, since it strictly reduces $W_{tot}$ and therefore raises the bound, while retaining them with an appropriate weight does not. In Section~\ref{sec:results}, we will test and demonstrate this effect numerically.
\end{Remark}

Conditions \eqref{eq:cond1}-\eqref{eq:cond2} characterize the optimum but do not indicate how to construct it. We now show that the Weighted-CVT of Section~\ref{subsec:voronoi} satisfies both asymptotically, as a structural consequence of the construction rather than through explicit optimization.

The bisector between robots $i$ and $j$ induced by the weighted distance of eq.~\eqref{eq:weighted_dist} is given by $\nu_i\|\mathbf{q}-\pos_i\|=\nu_j\|\mathbf{q}-\pos_j\|$. As a result, the region $\tilde V_i$ has a characteristic linear extent $\rho_i\propto\nu_i^{-1}$, and since $\mathcal{S}$ is a two-dimensional surface, its area obeys:
\begin{equation} \label{eq:area}
A_i \;\propto\; \rho_i^2 \;\propto\; \nu_i^{-2} \;=\; w_i \,.
\end{equation}
In other words, the Weighted-CVT allocates spherical area in proportion to sensor precision. From this, the two optimality conditions follow directly.

\begin{Prop} \label{prop:cvt}
Let $\{\textbf{p}^c_i,\tilde V_i\}$ be a Weighted-CVT on $\mathcal{S}$, so that each $\textbf{p}^c_i$ is the constrained centroid of $\tilde V_i$ and, by eq.~\eqref{eq:area}, $A_i = |\mathcal{S}|\,w_i/W_{tot}$ up to a discretization error that vanishes as the regions become small. Then, for any smooth $f$,
\begin{equation} \label{eq:riemann}
\sum_{i=1}^N w_i f(\textbf{p}^c_i) = \frac{W_{tot}}{|\mathcal{S}|}\sum_{i=1}^N A_i f(\textbf{p}^c_i) \;\approx\; \frac{W_{tot}}{|\mathcal{S}|}\int_{\mathcal{S}} f(\mathbf{q})\,d\mathbf{q}\,,
\end{equation}
since $A_i f(\textbf{p}^c_i)$ is the contribution of region $i$ to the sum of $f$ over $\mathcal{S}$ evaluated at its centroid. Applying eq.~\eqref{eq:riemann} to $f(\mathbf{q})=\mathbf{q}$ and to $f(\mathbf{q})=\mathbf{q}\mathbf{q}^\top$ gives, respectively,
\begin{equation} \label{eq:limits}
\sum_i w_i\,\textbf{p}^c_i \approx \frac{W_{tot}}{|\mathcal{S}|}\int_{\mathcal{S}}\mathbf{q}\,d\mathbf{q} = \mathbf{0}\,, \qquad
\sum_i w_i\,\textbf{p}^c_i{\textbf{p}^c_i}^\top \approx \frac{W_{tot}}{|\mathcal{S}|}\int_{\mathcal{S}}\mathbf{q}\mathbf{q}^\top d\mathbf{q} = \frac{W_{tot}}{3}\mathbf{I}_{3\times3}\,,
\end{equation}
which are exactly conditions \eqref{eq:cond1} and \eqref{eq:cond2}.
\end{Prop}

The two optimality conditions are thus obtained simultaneously without being imposed. The CVT makes each generator the centroid of its own region, the regions tile the sphere, and the weighted construction makes each region's area proportional to that robot's precision weight. We can also note that the exponent in the weighted distance of eq.~\eqref{eq:weighted_dist} is not arbitrary. Had we adopted $d_i(\mathbf{x},\mathbf{y})=\nu_i^s\|\mathbf{x}-\mathbf{y}\|$ on a $d$-dimensional domain, we would obtain $A_i\propto\nu_i^{-sd}$, and matching the required $A_i\propto\nu_i^{-2}$ forces $s=2/d$. On a spherical surface, $d=2$ and hence $s=1$, which is the choice made in eq.~\eqref{eq:weighted_dist}. The correspondence is therefore specific to formations supported on a surface.
In order to further validate these results, in Section~\ref{subsec:optimality_check}, we will verify numerically that the resulting formation lies within a few percent of the bound, and quantify the residual gap with respect to the bound of eq.~\eqref{eq:bound}.

Finally, it is important to recall that the bound of eq.~\eqref{eq:bound} concerns only the stochastic error induced by the measurement noise. The truncation term $\varphi(\pos_i,\C)$ of eq.~\eqref{eq:taylor} generates an additional deterministic bias that grows with the formation radius $D$ and with the curvature of the field, and that the weighting cannot remove. The overall accuracy of the estimate therefore results from a trade-off in $D$ between this bias and the variance of eq.~\eqref{eq:bound}, which decreases as $D^{-2}$. For more details, an analysis of this effect, for different multi-robot formations, can be found in \cite{Lara2019TRO}.

\subsubsection{Weighted Voronoi-based Multi-Robot Formation Control}\label{sec:WCVT_formation_control}

The previous section presented the constrained CVT formation generator that allows obtaining a weighted Voronoi-based 3D spherical formation by taking into account the heterogeneity of the sensors measuring the scalar field of interest. This formation generator provides the desired positions in the formation for each robot of the team. To control the multi-robot system, we use the resulting desired positions of the CVT on the sphere as references for a position tracking controller for the robots.

Let us now define $\mathbf{p}_i^c$ for $i=1,...,N$ the centroids corresponding to a constrained CVT on a sphere centered at the origin and of unitary radius. We express the desired position $\mathbf{r}_i^{ref}$ of robot $i$ in the Weighted-CVT-based Spherical formation with respect to the global frame as:
\begin{equation}\label{eq:Voronoi_formation}
    \pos_i^{ref}=\C+D \; \textbf{p}^c_{i}, \quad \forall i=1, \ldots, N
\end{equation}
where $\C$ and $D$ are the center and the radius of the formation.

All the robots know the formation center $\mathbf{c}$ and the formation radius $D$, and are able to compute the centroid positions $\mathbf{p}_i^c$ of the constrained CVT formation. Therefore, each robot can compute its desired position $\mathbf{r}_i^{ref}$. In this situation, we can design a control strategy to ensure that each robot's position $\mathbf{r}_i$ converges to its reference $\mathbf{r}_i^{ref}$.

The controller will depend on the robot's dynamics, and in the case of heterogeneous robots, each one may have a different controller. Let us consider the general case of time-invariant nonlinear systems, such that
\begin{align}
    \nonumber \dot{\mathbf{X}}_i&=f(\mathbf{X}_i,\mathbf{u}_i)\\
    \nonumber \mathbf{r}_i&=g(\mathbf{X}_i,\mathbf{u}_i)
\end{align}
where $\mathbf{X}_i$ represents the state of robot $i$, $\mathbf{u}_i$ its control input and $f(\cdot)$ and $g(\cdot)$ are two continuously differentiable functions. Taking into account the dynamic model of the robots, a position controller can be designed to guarantee correct tracking of the reference position, such that $\pos_i \rightarrow \pos_i^{ref}$, as presented in Section~\ref{subsec:new_sim} with a practical example of a team of nonlinear mini UAVs (Unmanned Aerial Vehicles).

The global control approach is composed of three connected algorithms. At each iteration, each robot computes:
\begin{enumerate}
    \item The gradient estimation $\widehat{\nabla}\sigma(\mathbf{c})$ at the current formation center position using eq.~\eqref{eq:WLS}. The position of the center is updated following the gradient ascent in eq.~\eqref{eq:gradient_ascent}, where $\epsilon$ encompasses the sampling time $\Delta t$ and the gradient ascent step.
    \item Its new reference position $\mathbf{r}_i^{ref}$ by using the updated center position and the centroid positions of the weighted CVT formation generator described in Section~\ref{subsec:voronoi}.
    \item The control input $\mathbf{u}_i$ to ensure position tracking, i.e., $\mathbf{r}_i \rightarrow \mathbf{r}_i^{ref}$.
\end{enumerate}

\begin{figure}[t]
\includegraphics[width=\columnwidth]{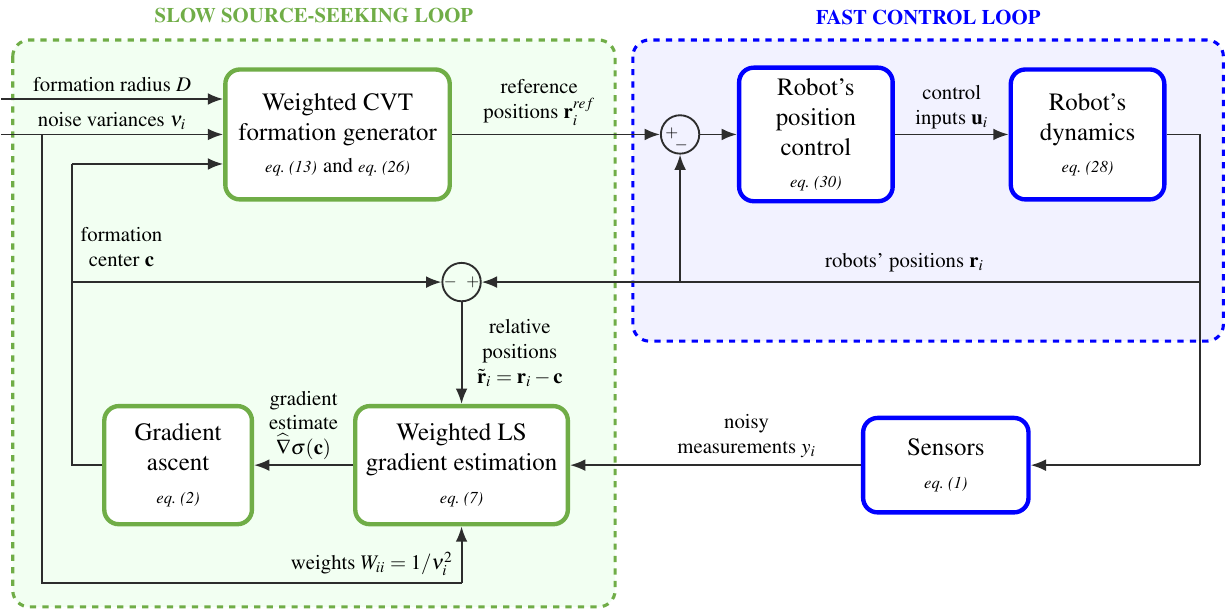}
\caption{Block diagram of the proposed source-seeking architecture. The inner (fast) loop stabilizes each robot on its reference position $\pos_i^{ref}$ within the Weighted-CVT formation, while the outer (slow) loop drives the formation center $\C$ along the estimated gradient. The sensor noise variances $\nu_i$ enter the architecture twice: as the weights of the Voronoi tessellation, which determine \emph{where} the measurements are collected, and as the weights of the WLS estimator, which determine \emph{how} they are combined.}
\label{fig:block_diagram}
\end{figure}

The interconnection of these three algorithms, together with the robots' dynamics and sensing, is summarized in Fig.~\ref{fig:block_diagram}. The architecture has a cascaded structure with two loops operating on separate time scales: an inner, fast loop stabilizes each robot on its reference position $\pos_i^{ref}$, while an outer, slow loop drives the formation center along the estimated gradient. The diagram also makes explicit the dual role played by the noise variances $\nu_i$, which parameterize both the formation generator and the estimator, and which constitutes the core idea of the proposed approach.

\begin{figure}[tb]
    \centering
\includegraphics[trim=3.5cm 9.5cm 3.5cm 11cm, clip=true,width=0.65\linewidth]{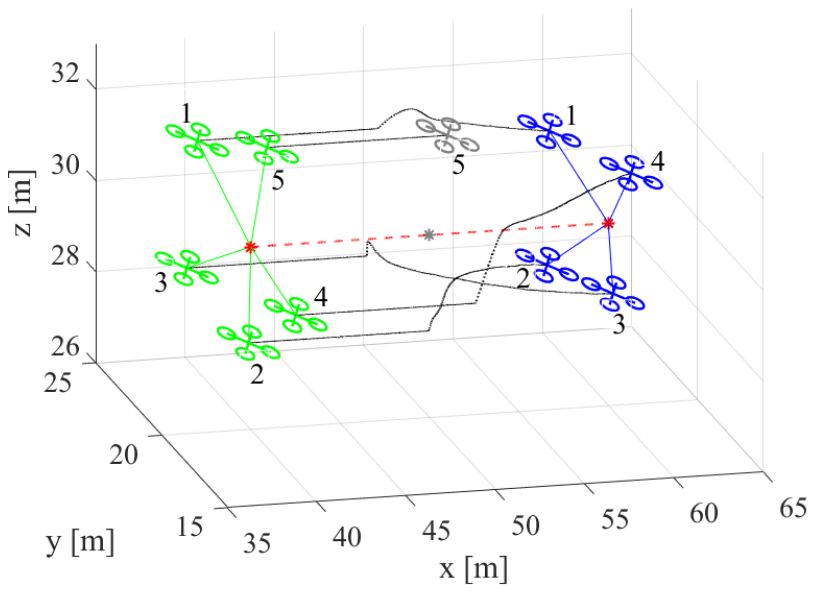}
    \caption{Reconfiguration of a weighted CVT formation when robot 5 fails. Green robots represent initial conditions at $t=0$. The formation center moves with a constant velocity of $0.1m/s$ along the x-axis. At $t=100s$, robot 5 fails (grey robot), and the CVT formation reconfigures online to obtain a new 4-robot weighted CVT formation, represented in blue. The red dashed line represents the center's trajectory and the black dotted lines the robots' trajectories.}
    \label{fig:reconfiguration}
\end{figure}

The advantage of this strategy is that the CVT formation can be computed online, and a reconfiguration is possible in the case of failure of one of the robots during the mission. The formation can reconfigure online by applying the Lloyd's algorithm. Fig.~\ref{fig:reconfiguration} illustrates a reconfiguration scenario where a moving Weighted-CVT-Spherical formation of 5 robots computes a new 4-robot formation when a robot fails and leaves the formation. In this scenario, we consider a heterogeneous team of $N=5$ robots where two of them have a sensor noise characterized by $\nu_i=0.8$ for $i=1,2$ while the remaining have $\nu_j=0.4$ for $j=3,4,5$. The robots compute their Weighted Voronoi regions online and converge to the new positions in the formation thanks to the position tracking controller, as shown by the robots' trajectories.

By presenting a unified strategy that simultaneously addresses formation control and gradient estimation via the optimal positioning of a heterogeneous team of robots in 3D, our proposed solution ensures both the maintenance of a desired formation and the enhancement of gradient estimation accuracy through strategic exploitation of sensor diversity and spatial configuration.

We emphasize that the framework proposed in this section is entirely model-based and does not employ any learning-based component. The gradient estimate is obtained in closed form, and the formation is the fixed point of a Lloyd iteration. As a result, it neither requires training data, an offline learning phase, nor any prior knowledge of the field beyond Assumption~\ref{assum:signal3D}. This choice has three important consequences. First, the estimator is unbiased for the linearized model and its covariance is available analytically, which is what makes the optimality analysis of Section~\ref{subsec:optimality} possible. Second, the method requires no field-specific tuning and therefore transfers between different fields without retraining. Third, reconfiguration after a robot failure is obtained by simply re-running the Lloyd iteration with $N-1$ generators, so that adaptation is a consequence of the geometry of the problem rather than of retraining or of an additional supervisory layer.

\section{Results}
\label{sec:results}
In this section, we present numerical results to test and analyze the performance of the proposed approach. First, we verify numerically the optimality analysis presented in Section~\ref{subsec:optimality} by directly evaluating the covariance of the gradient estimate for the proposed and uniform formations. Next, we present comparative results for a source-seeking application to analyze the formation center's convergence to the source. Finally, we show the implementation of the proposed approach on a team of simulated UAVs in the event of a mid-mission UAV failure to demonstrate reconfiguration capabilities.

\subsection{Verification of the optimality conditions} \label{subsec:optimality_check}

As a first study, we test numerically the analytical results provided in Section~\ref{subsec:optimality} to isolate the impact of the formation on the gradient estimation before analyzing its influence on the source-seeking problem. It is worth noting that the covariance of eq.~\eqref{eq:cov_grad} depends only on the formation geometry and on the noise variances, and not on the scalar field being measured. The analysis of this section is therefore independent of the test case introduced in Section~\ref{subsec:setup}, and assesses a property of the formation design itself rather than of a particular field.
To perform such analysis, we define the main performance indicator $\xi$ as the relative excess of $\mathrm{tr}\,\mathrm{Cov}(\widehat\nabla\sigma(\C))$, that is
the total variance of the gradient estimate over the lower bound of eq.~\eqref{eq:bound}. Given a configuration $\{\textbf{p}^c_i\}$ and a set of noise standard deviations $\{\nu_i\}$, the matrix $\mathbf{S}_w$ of eq.~\eqref{eq:cov_grad} is assembled and $\mathrm{tr}\,\mathrm{Cov}(\widehat\nabla\sigma(\C)) =
\mathrm{tr}\,\mathbf{S}_w^{-1}$ is evaluated exactly, without simulating the field or the measurement noise. The excess, expressed as a percentage, is then given by:
\begin{equation} \label{eq:excess}
\xi\;=\;
100\,\left(\frac{\mathrm{tr}\,\mathbf{S}_w^{-1}-9/(D^2W_{tot})}{9/(D^2W_{tot})}\right)
\end{equation}
where $\xi=0$ corresponds to a formation attaining the bound and conditions \eqref{eq:cond1}-\eqref{eq:cond2} being met exactly. Since $\mathbf{S}_w$ scales as $D^2W_{tot}$, both $D$ and $W_{tot}$ cancel in eq.~\eqref{eq:excess}: the excess depends only on the normalized precision profile $\{w_i/W_{tot}\}$ and on the geometry of the formation, and is unchanged if all the $\nu_i$ are multiplied by a common factor.

\begin{table}[t]
\centering
\begin{tabular}{lcccc}
\hline
Formation & $\mathrm{tr}\,\mathrm{Cov}(\widehat\nabla\sigma)$ & $\xi$ & $\|\bar{\textbf{p}}_w\|$ & $\mathrm{cond}(\mathbf{S}_w)$ \\
\hline
Uniform-CVT  & $0.0321$ & $+15.5\%$ $[8.5,35.3]$ & $0.1804$ & $2.16$ \\
Weighted-CVT & $0.0282$ & $\phantom{1}+1.4\%$ & $0.0174$ & $1.29$ \\
\hline
\end{tabular}
\caption{Verification of the optimality conditions for $N=8$, $D=3\,$m, three robots with $\nu=0.8$ and five with $\nu=0.4$, for the Uniform-CVT and the Weighted-CVT formations. The bound of eq.~\eqref{eq:bound} for this configuration is equal to $0.0278$.}\label{tab:optimality}
\end{table}

For uniform CVT-based formations, the excess also depends on which of the geometrically equivalent sites are occupied by the noisy robots, a degree of freedom that the Weighted-CVT does not have since it re-tessellates the sphere. Values for the uniform formation are consequently averaged over the assignments of the noise levels, with the observed range reported alongside.

Table~\ref{tab:optimality} reports the verification for the team used throughout this Section, that is $N=8$ robots, with three of them having $\nu=0.8$ and five having $\nu=0.4$, and $D=3\,$m. For this configuration, the bound of eq.~\eqref{eq:bound} equals $0.0278$. The Weighted-CVT lies within $1.4\%$ of the bound, whereas the Uniform-CVT is on average $15.5\%$ above it. The values for the Uniform-CVT are averaged over all $56$ possible combinations of the noise levels related to the robots' positions, and the range of the excess values is shown in brackets. The weighted centroid $\|\bar{\textbf{p}}_w\|$ is an order of magnitude smaller for the weighted formation and the condition number of $\mathbf{S}_w$ is significantly closer to unity, confirming that conditions \eqref{eq:cond1} and \eqref{eq:cond2} are both nearly satisfied.

\begin{table}[h!] \footnotesize
\centering
\begin{tabular}{lccccc}
\hline
$\nu_{high}/\nu_{low}$ & $1$ & $2$ & $3$ & $4$ & $6$ \\
\hline
Uniform CVT  & $+0.4\%$ & $+15.5\%$ & $+31.6\%$ & $+42.9\%$ & $+55.8\%$ \\
\phantom{Uniform CVT} \emph{range} & \emph{--} & \emph{$[8.5,35.3]$} & \emph{$[17.4,90.4]$} & \emph{$[22.5,142.5]$} & \emph{$[27.2,208.4]$} \\
Weighted CVT & $+0.4\%$ & $\phantom{1}+1.4\%$ & $\phantom{1}+2.0\%$ & $\phantom{1}+2.2\%$ & $\phantom{1}+2.0\%$ \\
\hline
\hline
$N$ (noisy $k$) & $8$ $(3)$ & $16$ $(6)$ & $24$ $(9)$ & $40$ $(15)$ & \\
\hline
Uniform CVT  & $+15.5\%$ & $+5.9\%$ & $+3.7\%$ & $+2.1\%$ & \\
\phantom{Uniform CVT} \emph{range} & \emph{$[8.5,35.3]$} & \emph{$[0.9,12.7]$} & \emph{$[0.6,8.9]$} & \emph{$[0.4,6.3]$} & \\
Weighted CVT & $\phantom{1}+1.4\%$ & $+0.1\%$ & $+0.01\%$ & $+0.05\%$ & \\
\hline
\end{tabular}
\caption{Excess of $\mathrm{tr}\,\mathrm{Cov}(\widehat\nabla\sigma)$ as in eq.~\eqref{eq:excess}. Top: sensor disparity, with three noisy robots out of $N=8$. Bottom: team size at a constant noisy fraction $3/8$ and $\nu_{high}/\nu_{low}=2$.
Uniform-CVT values are averaged over different assignments of the noisy levels to the robots, exhaustively for $N=8$ and over $200$ random assignments otherwise, with the range in brackets.}
\label{tab:sweep}
\end{table}

Table~\ref{tab:sweep} extends the comparison by considering different configurations. In the first block, the sensor disparity is increased while the profile is held fixed at three noisy robots out of eight. The uniform formation degrades rapidly, whereas the weighted one remains within a few percent of the bound. As expected, at $\nu_{high}/\nu_{low}=1$, the two coincide, since the Weighted-CVT degenerates exactly to the Uniform-CVT. The spread of the uniform values also increases considerably with the disparity. For instance, at $\nu_{high}/\nu_{low}=6$, the penalty obtained with a uniform formation ranges from $+27\%$ to $+218\%$ depending only on where the noisy robots are placed.

The second block varies the team size at a constant fraction of noisy robots of $3/8$, so that the normalized profile is held fixed and only $N$ changes. This is done to isolate the asymptotic behavior of Proposition~\ref{prop:cvt}. The excess $\xi$ of the Weighted-CVT falls below $0.1\%$ from $N=16$ onward, confirming that the approximation of eq.~\eqref{eq:riemann} becomes accurate quickly. The advantage over the uniform formation nevertheless decreases with $N$, since the weighted moments of any well-spread configuration progressively average out. The proposed weighting is thus most valuable in the small-team regime that motivates this work.

\subsection{Simulation setup} \label{subsec:setup}

\begin{figure}[t]
     \centering
    \includegraphics[trim=5.5cm 3.2cm 5.5cm 3cm, clip=true,width=1\columnwidth]{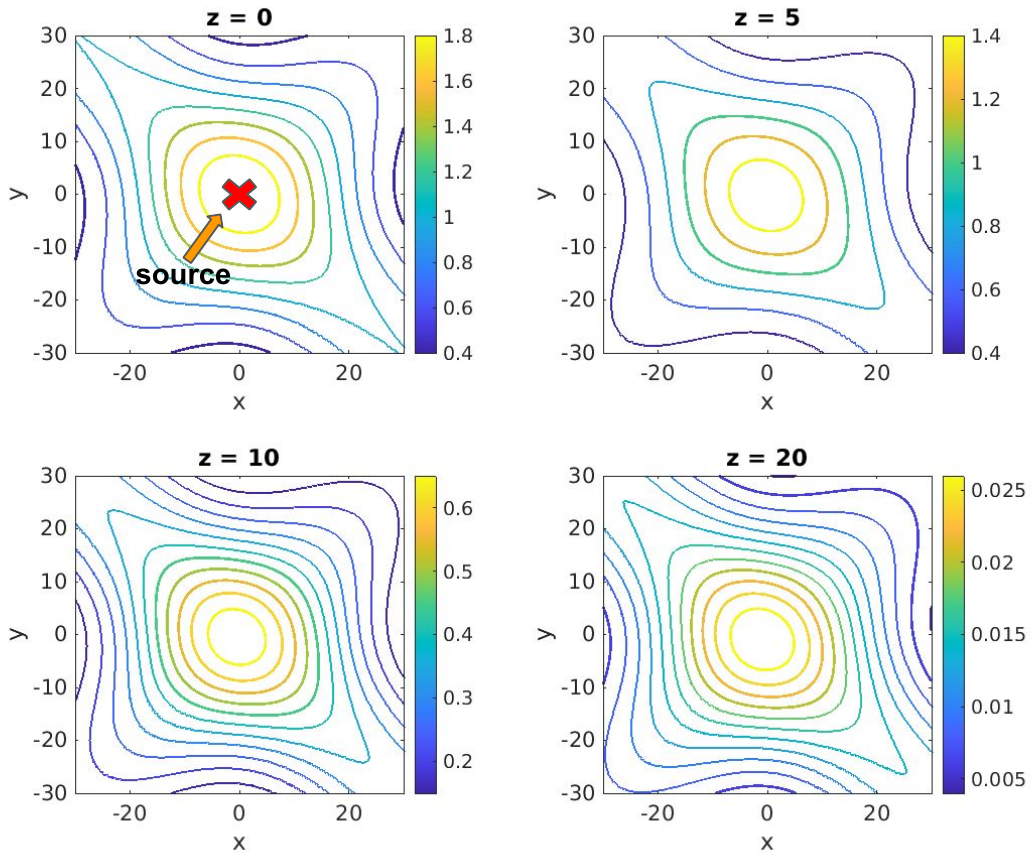}
        \caption{The 3D scalar signal field used in numerical simulations. The non-convex level curves in the $x-y$ plane are shown for 4 different values of altitude $z$. The robots have the goal of localizing the source based on noisy measurements of the field.}
        \label{fig:field}
\end{figure}

To model a generic physical signal propagating into 3D space, we use Gaussian functions. The use of Gaussian functions is a standard and commonly adopted choice in the source localization literature, as Gaussian fields can effectively approximate a broad range of scalar fields encountered in real-world applications. This modeling approach has been adopted in numerous prior works, including \cite{zhang2023pso, li2022multi, fabbiano2014source, coleman2023exploration, kahn2015global, lin2020distributed}, among others. However, to avoid overly simplistic scenarios, such as those involving isotropic signal fields with convex level curves, which are relatively easy to navigate, we decided to adopt a more challenging test case by combining two Gaussian functions with covariance matrices that generate non-convex level curves. This choice allows us to generate more complex fields, providing a more demanding and representative setting for evaluating the robustness and generality of our approach. The considered field is thus represented by:
$$
\sigma(\pos)=\alpha e^{-(\pos-\pos^*)^\top\textbf{S}_1(\pos-\pos^*)} +\beta e^{-(\pos-\pos^*)^\top\textbf{S}_2(\pos-\pos^*)}
$$
with its maximum corresponding to the source located at $\pos^*$.
In this work, $\alpha=\beta=1$ and the matrices $\mathbf{S}_{1,2}$ are set to:
\small
$$
\textbf{S}_1=10^{-4} \begin{bmatrix}
20 & 20 & 1\\
20 & 20 & 1\\
1 & 1 & 100\\
\end{bmatrix}
\quad \textbf{S}_2=10^{-4} \begin{bmatrix}
20 & -15 & 1\\
-15 & 20 & 1\\
1 & 1 & 100\\
\end{bmatrix}
$$
\normalsize
The values of $\textbf{S}_1$ and $\textbf{S}_2$ have been fixed to create a non-trivial, non-isotropic scalar field with non-convex level curves, shown in Fig.~\ref{fig:field}, able to model a wide range of physical signals and diffusion models.

The two amplitudes are set equal so that neither Gaussian component dominates the other: as $\alpha/\beta\to0$ or $\infty$, the superposition degenerates into a single anisotropic Gaussian with convex level curves, which is precisely the simplified case we wish to avoid. Their common magnitude, together with the noise levels $\nu_i$, then fixes the signal-to-noise ratio and hence the difficulty of the estimation problem, although by eq.~\eqref{eq:cov_grad} the estimation variance itself is independent of this scaling. The value $\alpha=\beta=1$ was chosen so that, at the initial formation center, this ratio produces a scenario with a demanding but non-degenerate regime.
The parameter $\epsilon$ defining the gradient ascent gain has been set to $\epsilon=5$, and the radius is $D=3$m.

\subsection{Convergence to the source} \label{subsec:convergence}

\begin{figure}[t]
     \centering
    \includegraphics[trim=0cm 2.5cm 0cm 1cm, clip=true,width=0.95\columnwidth]{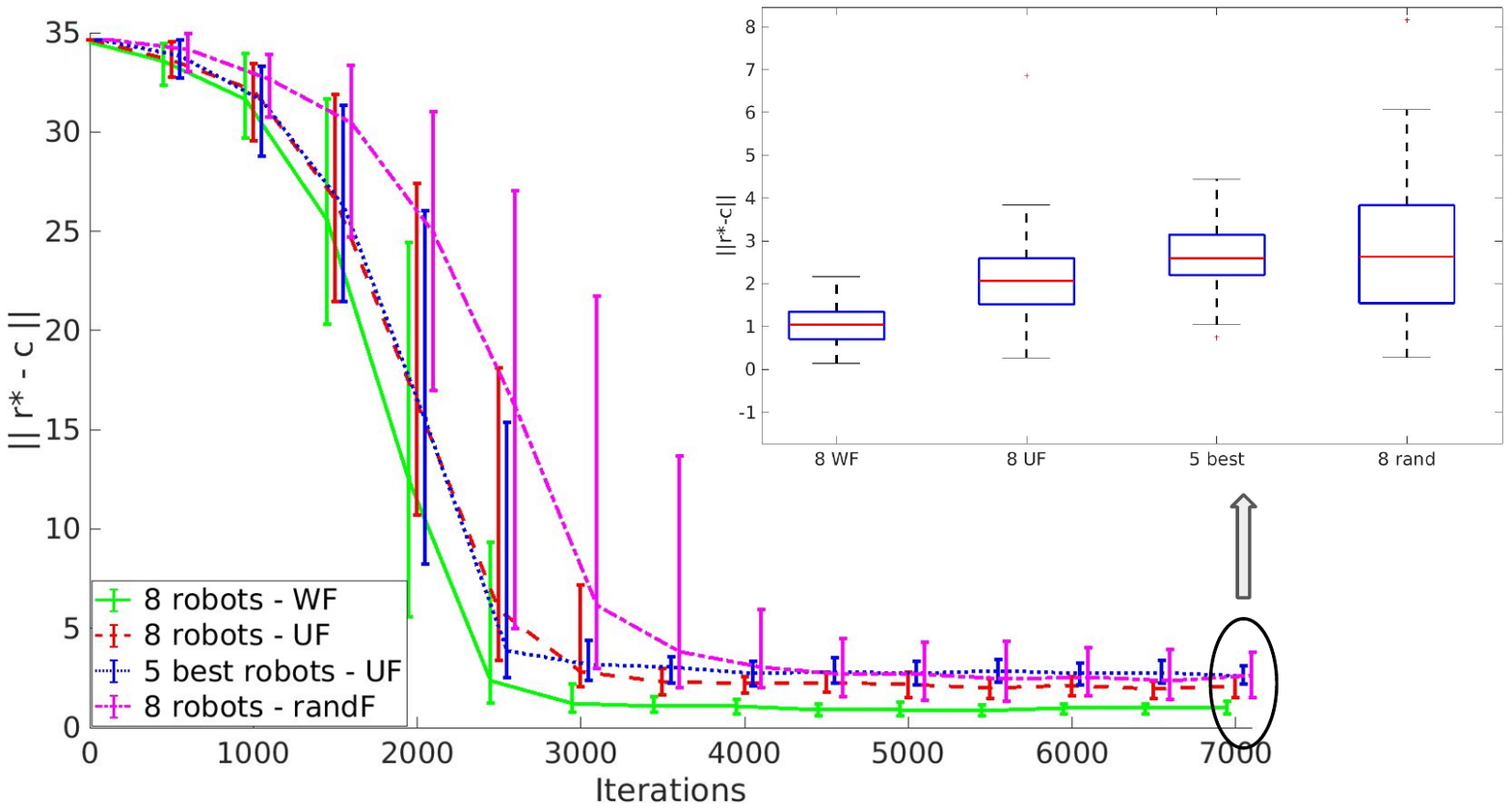}
        \caption{Evolution of the distance between the formation center and the source location over time. The values represent the median across 100 trials, with lower and upper bars indicating the 25th and 75th percentiles.}
        \label{fig:dist_comparison}
\end{figure}

The remainder of this section evaluates the proposed strategy in closed loop on a scalar field, where the multi-robot team follows the estimated gradient to localize its source.
In the first testing scenario, we consider a heterogeneous team of $N=8$ robots where five of them have sensor noise characterized by $\nu_i=0.4$ for $i=1,\cdots,5$, while the remaining three collect measurements with a higher level of noise, with $\nu_j=0.8$ for $j=6,7,8$. As a first test, we considered a fixed starting position for the center of the formation $\C(k=0)=[20,20,20]^\top$, and the source is located at $\pos^*\!=\![0, 0, 0]^\top$. To study the performance of the proposed strategy, we analyze the distance between the center and the location of the source over time. In particular, we compare the results obtained by the proposed non-uniform Weighted Formation (WF) with three different alternatives for the multi-robot formation:  {\it i)} the Uniform Formation (UF) obtained from a standard centroidal Voronoi partition of the spherical surface; {\it ii)} the same type of uniform formation but using only the five most accurate robots, to study the effect of the less precise measurements;  {\it iii)} a random formation (randF) where the robots' positions are generated randomly from a uniform distribution over a spherical surface. In all these cases, the gradient estimation is always obtained via weighted least squares, as in eq.~\eqref{eq:WLS}. The results of this analysis are shown in Fig.~\ref{fig:dist_comparison}, where the values in the plot represent the medians obtained across $100$ trials. We can notice that, even though all solutions are able to converge toward the source, the non-uniform weighted formation is both the fastest and the one obtaining the most accurate localization of the source, i.e., the smallest final error $||\pos^*-\C||^2$ and with smaller dispersion. In addition to the curve over time, a box plot helps to better visualize the data distribution of the last algorithm iteration.

\begin{figure}[t]
     \centering
    \includegraphics[trim=1cm 0cm 1cm 2cm, clip=true,width=.8\columnwidth]{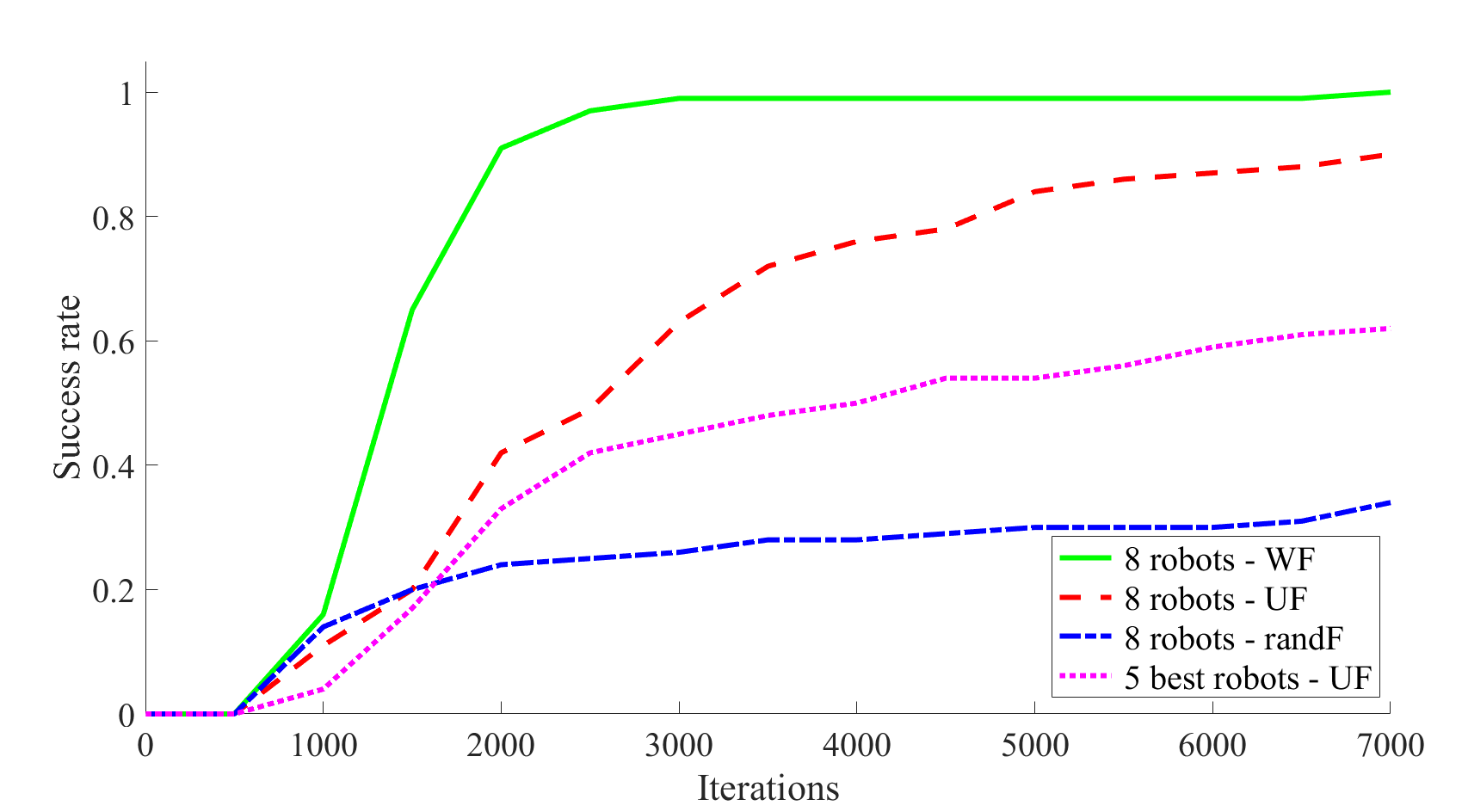}
        \caption{Success rate for the 4 different multi-robot formations corresponding to a distance threshold $\Delta=1.5\,$m.}
        \label{fig:succ_rate}
\end{figure}

To ensure that this effect is not specific to a given initial configuration, we then conduct a similar analysis but by varying the initial position of the formation center randomly at a fixed distance from the source $||\pos^*-\C(k=0)||=25\,$m. To compare the results obtained for this scenario, we study the success rate by defining a threshold distance $\Delta$ between the source and the center of the formation, below which the trial is considered successful. Fig.~\ref{fig:succ_rate} shows the results corresponding to $\Delta=1.5$m for the same four multi-robot formation strategies. Also in this case, we can see how the proposed non-uniform weighted formation significantly outperforms the other solutions by being the only one to reach $100\%$ of success. Interesting conclusions can also be drawn from the behavior of the other approaches. First of all, the importance of carefully designing the multi-robot formation is highlighted by the poor performance of the randomly designed formation. Additionally, it is interesting to notice that the lower weights assigned to the least accurate sensor do not simply have the effect of canceling out their contributions but are able to leverage them in the right way and clearly obtain better results than the formation composed uniquely of the most accurate sensors.

\subsection{Simulation results with a team of UAVs} \label{subsec:new_sim}

The goal of this section is to show how our source-seeking solution can be implemented to control a team of UAVs. To this end, we combine the source-seeking path planning strategy with a position controller to stabilize the robots in the desired Weighted-CVT-based formation and show the correct behavior of the multi-robot system. This implementation has been carried out in a realistic Matlab Simulink simulator developed at GIPSA-lab, which takes into account the inner-loop dynamics of the real Parrot Mambo UAVs, identified using flight test data \cite{Kostas_PhDThesis}. The Parrot Mambo is a quadrotor, i.e., a rigid body with six degrees of freedom, three translational and three rotational, actuated by four independent rotors. The simulator models the nonlinear dynamics of the mini-UAVs without taking into account phenomena such as gyroscopic effects, ground effects, blade flapping, and air friction, but considering the dynamics of the motors. A description of the simulator and the numerical values of the platform can be found in \cite[Section 2.4]{Kostas_PhDThesis}

It is worth emphasizing that the proposed framework does not depend on this particular platform, nor on aerial vehicles at all. The formation generator of Section~\ref{subsec:voronoi} produces reference positions $\pos_i^{ref}$, and the only requirement placed on the robots is that they can track these references with a settling time short compared to the motion of the formation center, as discussed in Section~\ref{sec:WCVT_formation_control}. The quadrotor model is adopted here to demonstrate feasibility on a realistic, underactuated and unstable platform, but other holonomic or non-holonomic vehicles equipped with a position-tracking controller could be substituted without modifying the estimation or formation-design layers. The LQR design presented below is also not essential. The inner loop may also be realized with nonlinear, adaptive, or intelligent control schemes, which are regularly adopted for multi-rotor vehicles \cite{lopez2023pid, Nascimento2019} and also applied to other nonlinear mechanical positioning systems \cite{bachi2021design}, provided that the resulting settling time remains short compared with the motion of the formation center.

We focus here on the position and velocity control of the UAVs, considering that the inner control loops (rotors' speeds and attitude control) are correctly designed. Let us define an inertial global frame described by the unit vectors $\{\textbf{e}_x, \textbf{e}_y, \textbf{e}_z\}$ and a body frame attached to each UAV's center of mass, moving along with its motion, with unit vectors $\{\textbf{b}_{xi},\textbf{b}_{yi}, \textbf{b}_{zi}\}$. Let us define the total thrust expressed in the body frame of the UAV $i$ as $T_i \textbf{b}_{zi}$ and the rotation matrix from the body to the inertial frame as $\textit{\textbf{R}}_i$. The translational dynamics of each robot $i$ can be expressed as:
\begin{equation} \label{eq:transl_dynamics}
    \begin{split}
        \dot{\pos}_i  &= \textbf{v}_i \\
        m \dot{\textbf{v}}_i &= -m g \textbf{e}_z +   \textit{\textbf{R}}_i T_i \textbf{b}_{zi}
    \end{split}
\end{equation}
where $\pos_i$ denotes the position of robot $i$, $\textbf{v}_i$ its velocity, $m$ its mass and $g$ the gravity.
The control inputs are the Euler angles and the magnitude of the total thrust, i.e., $\textbf{u}_i = [\phi_i, \theta_i, \psi_i, T_i]^\top$. Linearizing the translational dynamics around a hovering position, the UAV dynamics can be expressed as:
\begin{equation}\label{eq:UAV_state_space_model}
    \begin{bmatrix}
        \dot{\pos}_i\\
        \dot{\textbf{v}}_i
    \end{bmatrix} = \begin{bmatrix}
        \textbf{0}_{3\times3} & \textbf{I}_{3\times3}\\
        \textbf{0}_{3\times3} & \textbf{0}_{3\times3}
    \end{bmatrix}\begin{bmatrix}
        \dot{\pos}_i\\
        \dot{\textbf{v}}_i
    \end{bmatrix} + \begin{bmatrix}
        0 & 0 & 0 & 0  \\
        0 & g & 0 & 0  \\
        -g & 0 & 0 & 0 \\
        0 & 0 & 0 & 1/m\\
        \end{bmatrix}\textbf{u}_i
\end{equation}
where $\textbf{0}_{3\times3}$ and $\textbf{I}_{3\times3}$ denote, respectively, a matrix of zeros and the identity matrix of appropriate dimensions.

As presented in \cite{Kostas_ICUAS2024}, the UAVs can track their desired positions $\pos_i^{ref}$ with a linear state-feedback controller
\begin{equation}
    \textbf{u}_i = -K_P\left(\pos_i-\pos_i^{ref}\right) - K_D \textbf{v}_i - K_I \int_{0}^{t} \left(\pos_i-\pos_i^{ref}\right)dt
    \label{eq:LQR_integral_control}
\end{equation}
where the control gains have been tuned via an LQR method to obtain
\begin{equation*} \footnotesize
    \textbf{$K_P$}=\setstretch{1.2}
\setlength\arraycolsep{2pt}
  \begin{bmatrix}
        0 & -0.242 & 0 \\
        0.242 & 0 & 0 \\
        0 & 0 & 0.253
        \end{bmatrix},
        \quad
        \textbf{$K_D$}=\setstretch{1.2}
        \setlength\arraycolsep{2pt}
        \begin{bmatrix}
         0 & -0.244 & 0 \\
        0.244 & 0 & 0 \\
        0 & 0 & 0.218
        \end{bmatrix},
        \quad
        \textbf{$K_I$}=\setstretch{1.2}
        \setlength\arraycolsep{2pt}
        \begin{bmatrix}
        0 & 0.1 & 0 \\
        -0.1 & 0 & 0 \\
        0 & 0 & 0.1
        \end{bmatrix}.
\end{equation*}

With a view to stabilizing the team of UAVs to the desired formation, the desired positions are defined by eq.~\eqref{eq:Voronoi_formation}. Each robot knows the formation center $\C$ and radius $D$ and is able to compute its centroid position $\textbf{p}_i^c$ of the constrained CVT.

\begin{figure}[t]
     \centering
    \includegraphics[trim=0cm 0cm 0cm 0cm, clip=true,width=.85\columnwidth]{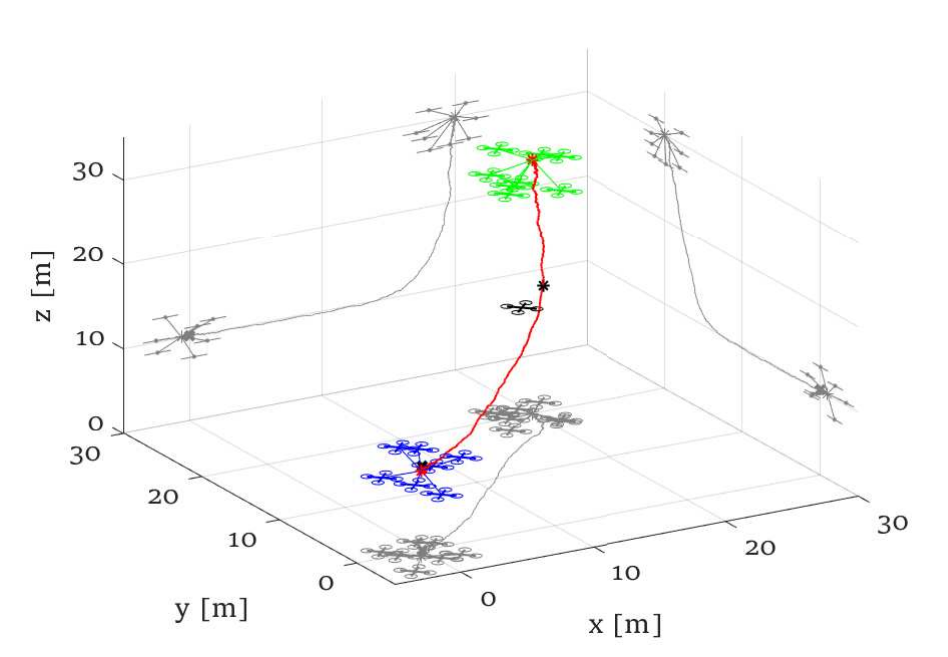}
        \caption{Simulation of a team of 8 Parrot Mambo UAVs with heterogeneous sensors performing a source-seeking mission. The trajectory of the formation center is in red. The 8-robot initial formation is in green. One robot (in black) fails at $t=300s$, and the final configuration of the 7-robot formation is shown in blue near the source location.}
        \label{fig:3D_Voronoi_source_seeking}
\end{figure}

Fig. \ref{fig:3D_Voronoi_source_seeking} illustrates a realistic simulation with a team of $N=8$ Mambo quadrotors equipped with heterogeneous sensors as in the first scenario presented previously, performing the proposed source-seeking strategy. The sensor noise for robots $i = 1, \ldots, 5$ is characterized by $\nu_i = 0.4$, while $\nu_j = 0.8$ for $j = 6, 7, 8$. The sampling time between two iterations is $0.01$s. In order to produce a realistic scenario in which the UAVs should not touch the ground, the source is located at $\pos^*\!=\![0, 0, 10]^\top$ and consequently the formation center starts at $\C(0)=[20,20,30]^\top$ to have a fair comparison with previous tests. To demonstrate the adaptability of the proposed strategy, at time $t=300$s, robot $8$ fails and leaves the formation. The rest of the robots in the team compute the new Voronoi-based formation online, now with $N=7$, while simultaneously computing the gradient estimate. Thanks to the separation scale principle, even if the estimate of the gradient has a greater error during the transition and thus the center of the formation moves slowly following an incorrect direction, the robots converge to the new desired formation, and the gradient estimate becomes precise enough to steer the center to the source. This highlights the capability of the proposed approach to handle unexpected failures online and achieve a correct reconfiguration of the formation.

\section {Discussion}

This section discusses the implications of the proposed approach, analyzes the observed performance, and highlights its limitations and potential extensions.

\subsection{Impact of Heteroskedastic Measurements on Gradient Estimation}
The simulation results presented in the previous section confirm that explicitly accounting for heteroskedastic measurement noise significantly improves the accuracy and robustness of gradient estimation. By weighting both the formation design and the estimation process according to sensor-dependent noise variances, the proposed approach prioritizes reliable measurements while reducing the influence of noisier sensors. This effect is particularly evident when comparing the proposed strategy with unweighted or homogeneous formations, where estimation accuracy degrades noticeably in the presence of heterogeneous sensing capabilities. In particular, we showed how the gap to the uniform formation increases with sensor disparity (Table~\ref{tab:sweep}) and is most pronounced for small teams, while it decreases as the team grows, since the weighted moments of any well-distributed large formation progressively average out.

These observations support the central premise of this paper: in multi-robot source-seeking problems, estimation quality and robot deployment must be jointly considered rather than treated as independent components. The Weighted-CVT provides an adaptive mechanism to couple estimation objectives with formation control. Unlike fixed or heuristic formations, the proposed method adapts the spatial configuration of the robots to the statistical properties of the sensors, resulting in measurement locations that are better suited for gradient estimation.
More specifically, as shown in Section~\ref{subsec:optimality}, the resulting formations satisfy the optimality conditions derived in Proposition~\ref{prop:bound} as a structural consequence of their construction, and Section~\ref{subsec:optimality_check} confirmed numerically that they remain within a few percent above the theoretical lower bound of eq.~\eqref{eq:bound}, against excess values of above $15\%$ for the uniform formation under the same conditions (Table~\ref{tab:optimality}).
These considerations can also be applied to different cooperative, sensing-based, multi-robot strategies. Although noise models are typically used to design improved estimation approaches, they are rarely leveraged explicitly for the coordination and control of robots.

\subsection{Reconfiguration and Adaptability to Robot Failures}
One of the key strengths of the proposed framework is its intrinsic ability to reconfigure online in response to changes in team composition, such as robot failures. When a robot becomes unavailable, the formation automatically adapts through the same weighted CVT mechanism without requiring explicit fault detection, task reassignment, or redesign of the control architecture.

The presented failure scenarios demonstrate that, after a transient phase, the formation converges to a new equilibrium configuration and the gradient estimation accuracy recovers sufficiently to resume effective source-seeking behavior. This highlights an important property of the approach: the ability to adapt emerges naturally from the formulation rather than being enforced through additional supervisory layers.

\subsection{Assumptions on Sensor Noise Modeling}
In conclusion, it is important to note that the proposed framework assumes the noise variance associated with each sensor is known a priori and remains constant during operation. Such variances can be obtained through prior calibration or manufacturer specifications.
However, sensor noise characteristics may evolve over time due to environmental conditions, hardware degradation, or partial sensor failures. In such cases, a mismatch between the assumed and actual noise variances may reduce the effectiveness of the weighting strategy and lead to suboptimal formation configurations or biased gradient estimates.

Therefore, an important extension of this work would be to incorporate online estimation or adaptation of the sensor noise variances. For instance, residual-based approaches could be employed to detect inconsistencies between predicted and observed measurements, enabling the weights used in the formation and estimation stages to be updated online. Such mechanisms would allow the framework to adapt not only to changes in team composition but also to evolving sensor reliability.

\section{Conclusions}
\label{sec:conclusion}

This paper presented a novel solution for locating the source of a signal propagating in a three-dimensional environment using a heterogeneous team of robots equipped with signal strength sensors affected by Gaussian noise with different variances. The proposed multi-robot formation strategy leverages a centroidal Voronoi partition based on distances weighted by measurement noise, allowing the system to self-organize into a non-uniform configuration that incorporates sensor heterogeneity. To demonstrate the advantages of these formations, we derived a closed-form expression for the covariance of the resulting gradient estimate, together with a lower bound, valid for any formation on the sphere, on the total variance of that estimate. We then showed that the proposed Weighted-CVT formation satisfies the associated optimality conditions up to an error that vanishes with the team size, a result then confirmed numerically even for small teams. Data collected from this optimized formation are then fed into a Weighted Least-Squares (WLS) estimation method, resulting in a robust and reliable estimation of the signal gradient at the center of the formation, which is then used to guide the robots toward the source.

Simulation results demonstrated that the proposed approach successfully drives the multi-robot system to locate the source of the field. Furthermore, a comparative analysis showed how this solution significantly outperforms alternative multi-robot formation strategies in both convergence speed and accuracy. We finally demonstrate the applicability of our solution for robotics systems by presenting a realistic simulation incorporating the dynamic model and a controller designed specifically for a group of UAVs. These results also showed the intrinsic ability of the proposed formation control strategy to reconfigure online in response to unforeseen situations.

For future work, we aim to improve our solution by incorporating a strategy that recursively estimates the confidence weights in the WLS problem to indirectly capture variations in the sensor variances, which could be caused by sensor deterioration, malfunction, or model errors. This would also trigger dynamic reconfigurations of the multi-robot formation, as in the case of a robot's failure shown in Section \ref{subsec:new_sim}.

\bibliographystyle{elsarticle-num}
\bibliography{biblio}

\end{document}